\documentclass[10pt,twocolumn,letterpaper]{article}

\usepackage{cvpr}              % To produce the CAMERA-READY version
\usepackage{multirow}
\definecolor{cvprblue}{rgb}{0.21,0.49,0.74}
\usepackage[pagebackref,breaklinks,colorlinks,allcolors=cvprblue]{hyperref}
\usepackage{pifont} 
\usepackage{fancyvrb}
\usepackage{xcolor}
\usepackage[most]{tcolorbox}
\usepackage{fancyvrb} % 支持自动换行的代码环境
\usepackage{fancyvrb} % 加载fancyvrb
\usepackage{listings}

\definecolor{codegray}{gray}{0.95}
\def\paperID{20530} % *** Enter the Paper ID here
\def\confName{CVPR}
\def\confYear{2026}
\newcommand{\MODELNAME}{IMAgent}
\newcommand{\DATASET}{MIFG-QA}
\title{\MODELNAME: Training Multi-Image Vision Agents via End2End Reinforcement Learning}

\author{
Chengqi Dong\textsuperscript{\rm 1,2},
Chuhuai Yue\textsuperscript{\rm 1},
Hang He\textsuperscript{\rm 1}, 
Rongge Mao\textsuperscript{\rm 2}, 
Fenghe Tang\textsuperscript{\rm 2}, 
S Kevin Zhou\textsuperscript{\rm 2}, \\
Zekun Xu\textsuperscript{\rm 1}, 
Xiaohan Wang\textsuperscript{\rm 1}, 
Jiajun Chai\textsuperscript{\rm 1}, 
Guojun Yin\textsuperscript{\rm 1} \\
\small\textit{\textsuperscript{1} MeiTuan, \textsuperscript{2} University of Science and Technology of China}
}
\begin{document}

\maketitle
\begin{abstract}

Recent VLM-based agents aim to replicate OpenAI O3's "thinking with images" via tool use, yet most open-source methods restrict inputs to a single image, limiting their applicability to real-world multi-image QA tasks. To address this gap, we propose \MODELNAME, an open-source visual agent trained with end-to-end reinforcement learning for fine-grained single/multi-image reasoning. During inference, VLMs tend to gradually neglect visual inputs; to mitigate this issue, we design two dedicated tools for visual reflection and confirmation, enabling the model to actively refocus attention on image content. Beyond that, we, for the first time, reveal how tool usage enhances agent performance from an attention perspective. Equipped with a carefully designed two-layer motion trajectory masking strategy and tool-use reward gain, \MODELNAME~acquires an effective tool-use paradigm through pure reinforcement learning, eliminating the need for costly supervised fine-tuning data.
To further unleash the inherent tool-usage potential of the base VLM and fill data gaps, we construct a challenging, visually enriched multi-image QA dataset via multi-agent system.
Extensive experiments validate that \MODELNAME~achieves SOTA performance across mainstream single and multi-image benchmarks, and our in-depth analysis offers actionable insights for the community. Code and data will be released soon.

\end{abstract}

\section{Introduction}

\begin{figure}[htbp]
  \centering
   \includegraphics[width=0.98\linewidth]{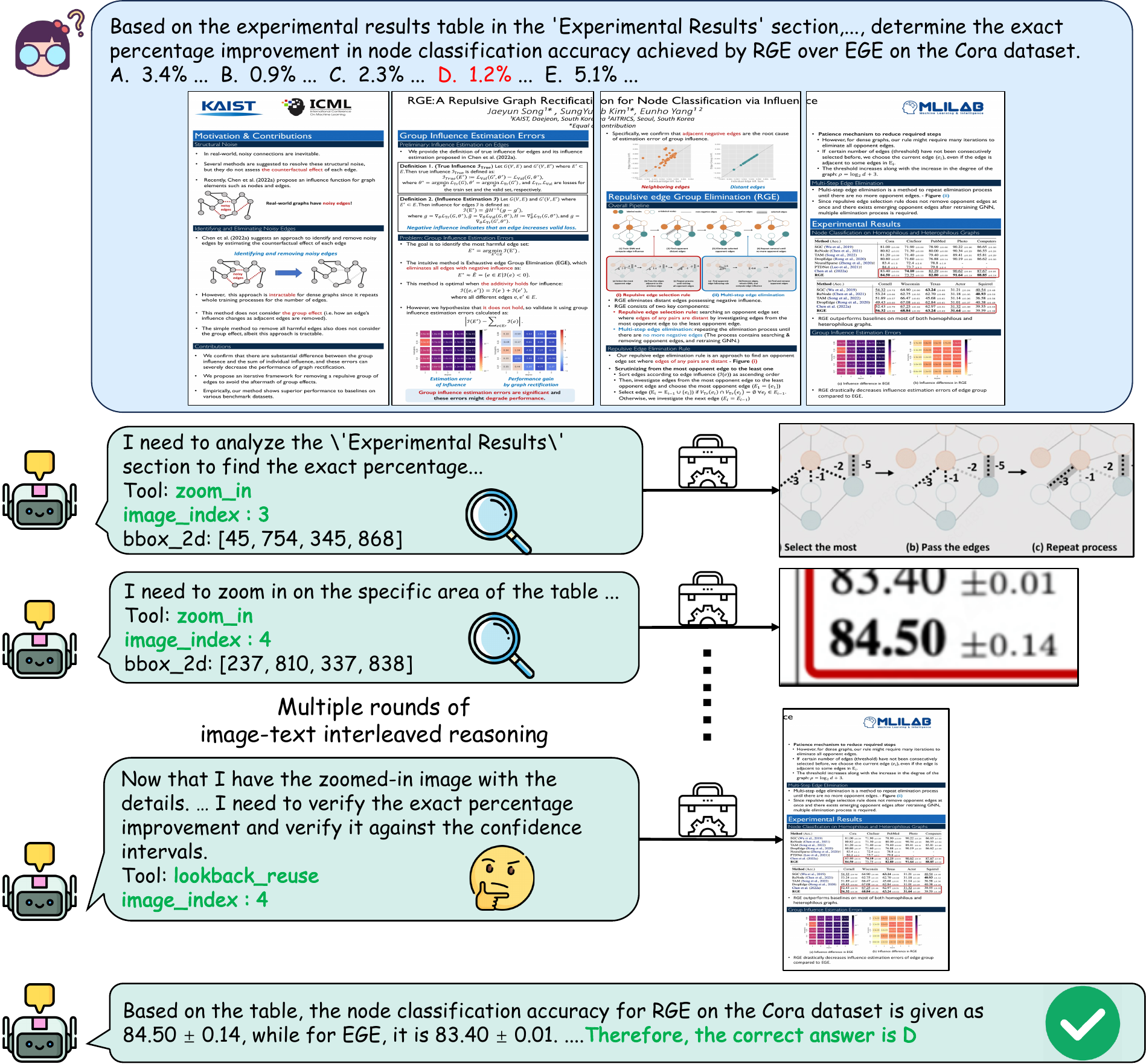}
   \caption{IMAgent flexibly select target image from multiple inputs for confirmation and reflection.}
   \label{fig:shoutu}
\end{figure}

\begin{figure*}[t]
  \centering
   \includegraphics[width=0.98\linewidth]{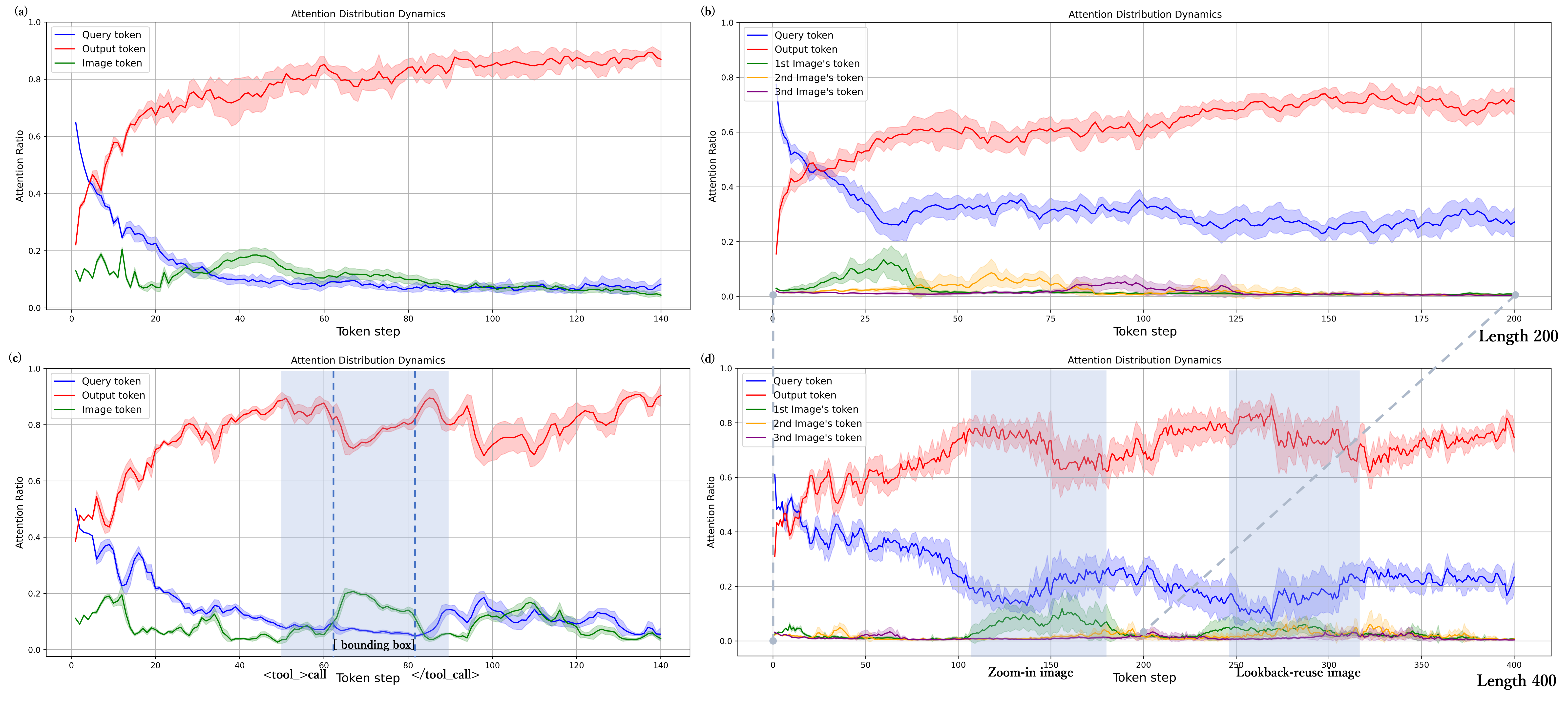}
   \caption{We compared the attention proportions by Qwen2.5-VL-7B (a, b) and \MODELNAME~ (c, d) to the input text, image tokens, and previously generated tokens as inference progresses, under single-image input (a, c) and multi-image input (b, d). Qwen2.5-VL-7B’s attention to image tokens gradually decreases and approaches zero under multi-image input (b); in contrast, \MODELNAME~ (c, d), aided by tool use, maintains dynamic visual attention throughout, even with a reasoning chain twice as long.}
   \label{fig:attention-decay}
\end{figure*}

The success of agentic reinforcement learning (RL)~\cite{jiang2024mmsearch,jin2025search,singh2025agentic} is driving vision-language models (VLMs) beyond passive perception and giving rise to \emph{Vision Agents}~\cite{acharya2025agentic,zhang2025landscape}. OpenAI's O3~\cite{OpenAIo3} demonstrates that agents can autonomously invoke visual tools to achieve ``thinking with images.'' Inspired by this, the open-source community trains tool-augmented agents (e.g., using zoom\_in to extract local details) that already achieve strong progress in single-image fine-grained scenarios~\cite{fu2025refocus,zhang2025chain,zheng2025deepeyes,wu2025vtool,su2025pixel}.

However, real-world applications widely require fine-grained understanding across multiple images~\cite{ye2025m4bench,liu2024mibench}, where models must capture subtle local cues and perform cross-image comparisons to reach correct conclusions. Due to architectural and cognitive mismatches, naively transferring single-image agents to multi-image tasks often causes systematic failures.

At the operational level, existing agents typically assume a \emph{Monolithic Visual Context} and lack \emph{Explicit Visual Indexing}~\cite{zheng2025deepeyes,wu2025vtool,cao2025ground,zhang2025chain,su2025openthinkimg}, making it difficult to specify which visual entity should be manipulated in multi-image settings. At the cognitive level, long-chain reasoning naturally causes models to forget the initial visual input as text generation grows. Intensive inspection of local details in one image further consumes the context window, amplifying neglect of other images and breaking cross-image reasoning chains~\cite{tian2025identifying}.

To overcome this bottleneck, we propose \MODELNAME, the first open-source vision agent trained end-to-end with RL for multi-images environments. Architecturally, as shown in Figure~\ref{fig:shoutu}, we expand the action space to support explicit visual indexing, enabling robust cross-image understanding. Cognitively, inspired by visual forgetting, we equip \MODELNAME~with a collaborative visual function suite: visual confirmation (zoom\_in for local details) and visual reflection (lookback\_reuse to restore global context). Empirical analysis reveals that, without active tool use during chain-of-thought (CoT) reasoning, the model’s attention to input images gradually diminishes, especially in multi-image scenarios. As shown in Figure~\ref{fig:attention-decay}, token-level focus on image content drops to near zero in later inference stages. But \MODELNAME, through timely tool calls, refocuses image attention. Our pure RL framework allows the model to learn when, how, and which image index to interact with, establishing a robust ``observe--confirm/reflect--iterate'' paradigm.

Eliciting multi-turn tool use in multi-image settings is non-trivial. Single-image agents can often rely on supervised fine-tuning (SFT) to acquire reasoning patterns, but in an expanded action space, SFT leads to severe policy anchoring and high costs. To avoid blind exploration and to regularize the exploding multimodal action space for stable pure RL post-training, we introduce a Two-Level Mask strategy (action-level and trajectory-level), together with a correctness-coupled incentive mechanism.

Furthermore, existing work mainly focuses on using tools to improve model performance, either through SFT using carefully curated trajectory data~\cite{wang2025simple}, modifying RL algorithms~\cite{cao2025ground,zheng2025deepeyes,wu2025vtool}, or combining these two methods~\cite{su2025openthinkimg,su2025pixel,zhang2025chain}. Few studies explore the underlying principles by which tool use actually enhances agent performance. Based on the visual forgetting phenomenon in CoT, we conduct quantitative and qualitative analysis from an attention perspective, revealing for the first time that tool use can help agents focus on task-relevant regions in a timely manner, thereby improving performance.

To fill the data gap, we curate \DATASET, a 10k-sample fine-grained multi-image dataset  for tool-interaction elicitation. Extensive experiments show \MODELNAME~maintains competitive single-image performance while significantly improving multi-image fine-grained reasoning, and demonstrates encouraging generalization on other multi-image benchmarks. Beyond accuracy, analyses show that the reasoning behavior after training repeatedly bring the model back to task-relevant visual evidence, validating the underlying mechanism.
\begin{figure*}[htbp]
  \centering
   \includegraphics[width=0.98\linewidth]{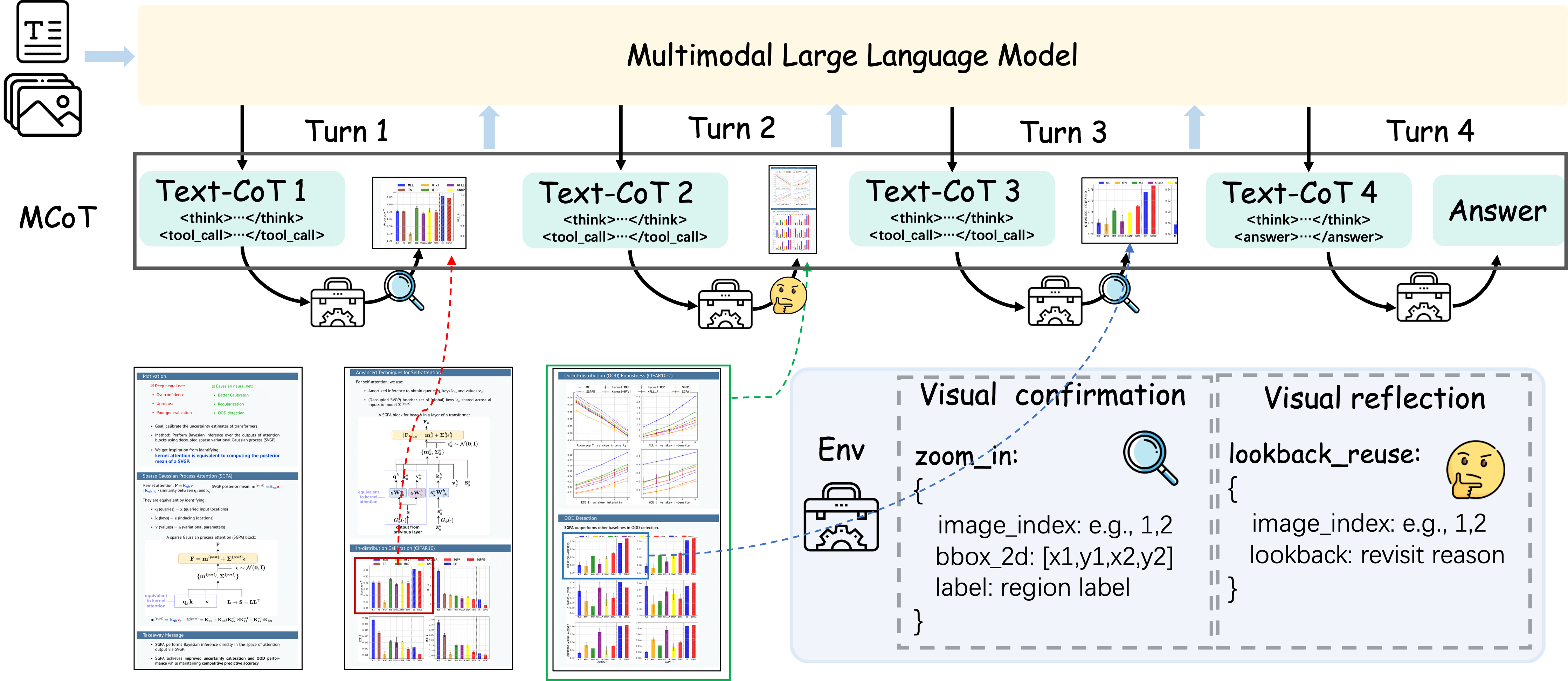}
   \caption{Overview of \MODELNAME. Our model will automatically choose whether and how to use tools based on the actual problem. Visual confirmation is achieved by cropping local regions of the image using zoom\_in, and visual reflection is achieved by reusing the input image using lookbask\_reuse.}
   \label{fig:model}
\end{figure*}
In summary, our main contributions are threefold:
\begin{itemize}
\item We propose \MODELNAME, the first multi-image, fine-grained, long-chain reasoning agent trained via end-to-end agentic RL, achieving dynamic cross-image coordination.
\item We identify systematic deficiencies in multi-image scenarios, specifically visual forgetting and the lack of explicit visual indexing.  To address this, We design collaborative visual tools and a two-level mask strategy, and firstly reveal how tool use influences changes in visual attention.
\item We construct the \DATASET~dataset to bridge the gap in multi-image fine-grained datasets. Extensive experiments on both single and multi-image datasets validate the state-of-the-art (SOTA) performance of \MODELNAME.
\end{itemize}

\vspace{-4pt}
\section{Related Work}

\subsection{Multi-modal Reasoning Models}

Multimodal reasoning models show a promising trend toward unifying multiple modalities within a language-centric framework\cite{li2025perceptionreasonthinkplan}. Early works\cite{xu2025llavacotletvisionlanguage,xu2025redstardoesscalinglongcot,yao2024mulberryempoweringmllmo1like} focus on enhancing CoT reasoning through SFT and planning algorithms like Monte Carlo Tree Search\cite{xie2024montecarlotreesearch}, improving interpretability and reliability. Building on these foundations, RL-based methods shift the paradigm from imitating static reasoning patterns to actively optimizing them via feedback. Pioneering methods\cite{yu2024rlhfvtrustworthymllmsbehavior,zhang2024improvevisionlanguagemodel,dong2025insightvexploringlongchainvisual} apply Direct Preference Optimization\cite{rafailov2024directpreferenceoptimizationlanguage} for behavioral alignment. After the success of Deepseek-R1\cite{deepseekai2025deepseekr1incentivizingreasoningcapability}, Group Relative Policy Optimization (GRPO)\cite{shao2024deepseekmathpushinglimitsmathematical} becomes a mainstream algorithm, and Reinforcement Learning with Verifiable Rewards (RLVR) emerges as a powerful new training paradigm. These methods rapidly expand from mathematical and geometric reasoning\cite{meng2025mmeurekaexploringfrontiersmultimodal,zhan2025visionr1evolvinghumanfreealignment,zhou2025r1zerosahamomentvisual} to visual skills\cite{liu2025visualrftvisualreinforcementfinetuning,liu2025segzeroreasoningchainguidedsegmentation,wang2025vicritverifiablereinforcementlearning}, and further to modalities such as video\cite{feng2025videor1reinforcingvideoreasoning} and audio\cite{zhao2025r1omniexplainableomnimultimodalemotion}. Their success highlights the importance of high-quality reward signals, making data augmentation\cite{wu2025synthrlscalingvisualreasoning,liang2025modomodomultidomaindatamixtures}, reward design\cite{zhang2025r1rewardtrainingmultimodalreward,wang2025unifiedmultimodalchainofthoughtreward,xu2025mixedr1unifiedrewardperspective,zhang2025viperempoweringselfevolutionvisual} and other modifications\cite{yue2025promotingefficientreasoningverifiable,guo2025ssl4rlrevisitingselfsupervisedlearning} central to performance. Ultimately, RLVR drives breakthroughs in reasoning depth, coherence, and domain adaptability, laying a foundation for training advanced vision agents.

\subsection{Vision Agents}

Vision Agents aim to break the limitations of static, parameterized knowledge by autonomously using tools, deeply integrating dynamic visual perception with action-based reasoning\cite{chai2025rlfactoryplugandplayreinforcementlearning}. Early approaches relied on prompt engineering to passively trigger tools\cite{jiang2024mmsearchbenchmarkingpotentiallarge,wang2025mllmtoolmultimodallargelanguage,gao2025multimodalagenttuningbuilding,yue-etal-2025-uiorchestra}, while current mainstream methods increasingly adopt RL post-training, driving advances in multimodal retrieval\cite{wu2025mmsearchr1incentivizinglmmssearch,geng2025webwatcherbreakingnewfrontier,wang2025vragrlempowervisionperceptionbasedrag} and integrated code execution\cite{gou2024toratoolintegratedreasoningagent,wang2023mathcoderseamlesscodeintegration,li2025torlscalingtoolintegratedrl,feng2025retoolreinforcementlearningstrategic}. Notably, OpenAI's O3 and Qwen3-VL\cite{QwenTeam2025QwenVL} enable models to ``think with images", generating intertwined chains of thought that combine text and visuals. Inspired by O3's great performance, many open-source solutions have emerged\cite{zheng2025deepeyes,cao2025ground,zhang2025chain,lai2025minio3scalingreasoningpatterns,su2025pixel,zhang2025chainoffocusadaptivevisualsearch}, typically adopting a two-stage post-training process: SFT for tool call format learning, followed by RL to enhance visual tool use for complex tasks. However, this approach requires carefully curated SFT datasets and limits generalization. Deepeyes\cite{zheng2025deepeyes} cultivates tool-calling ability through pure reinforcement learning, but is limited to single image input and a single tool type. Though, aforementioned works achieve partial power of O3, they mainly focus on reproducing the tool use ability in single-image scenarios, largely neglecting the more common multi-image input settings. Addressing this gap is one of the main motivations of our work.

\begin{figure}[htbp]
  \centering
   \includegraphics[width=1.0\linewidth]{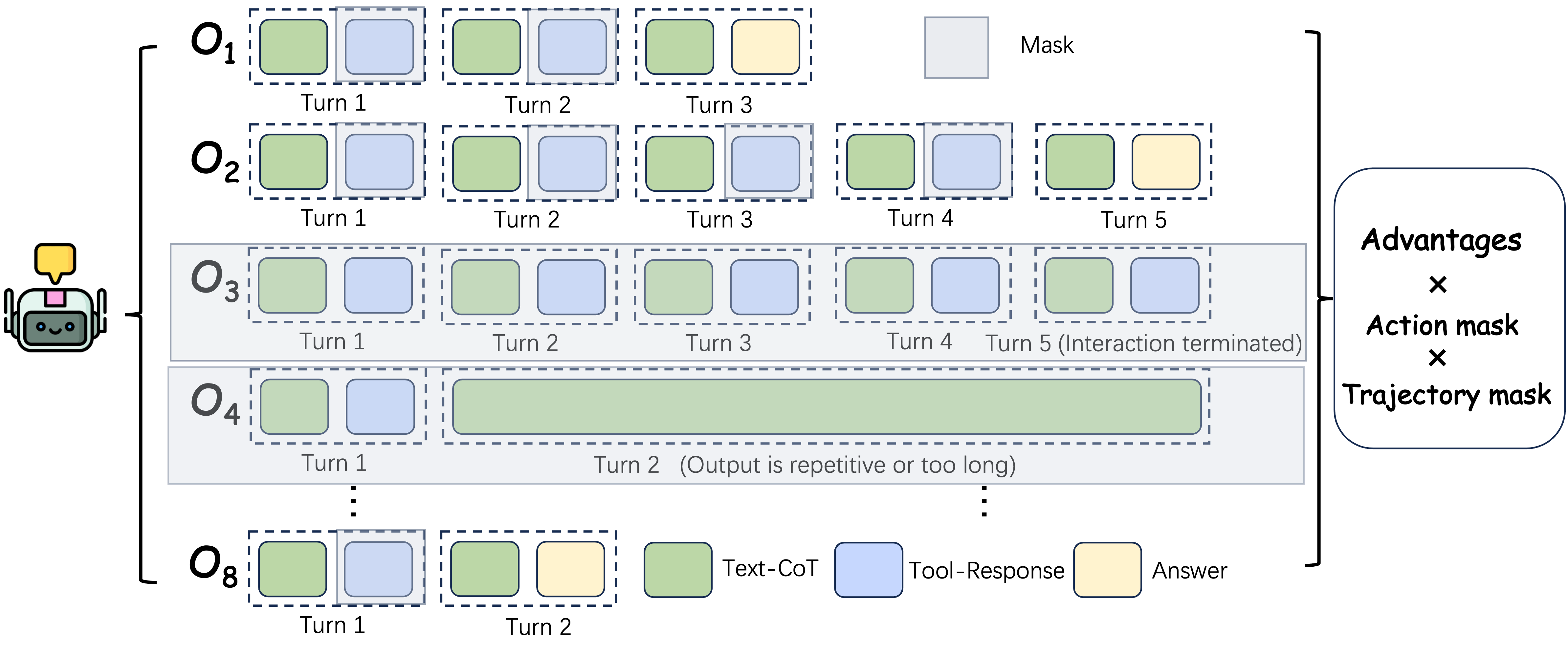}
   \caption{Action-Level and Trajectory-Level Mask.}
   \label{fig:twomask}
\end{figure}

\section{Method}

\begin{figure*}[htbp]
  \centering
   \includegraphics[width=0.98\linewidth]{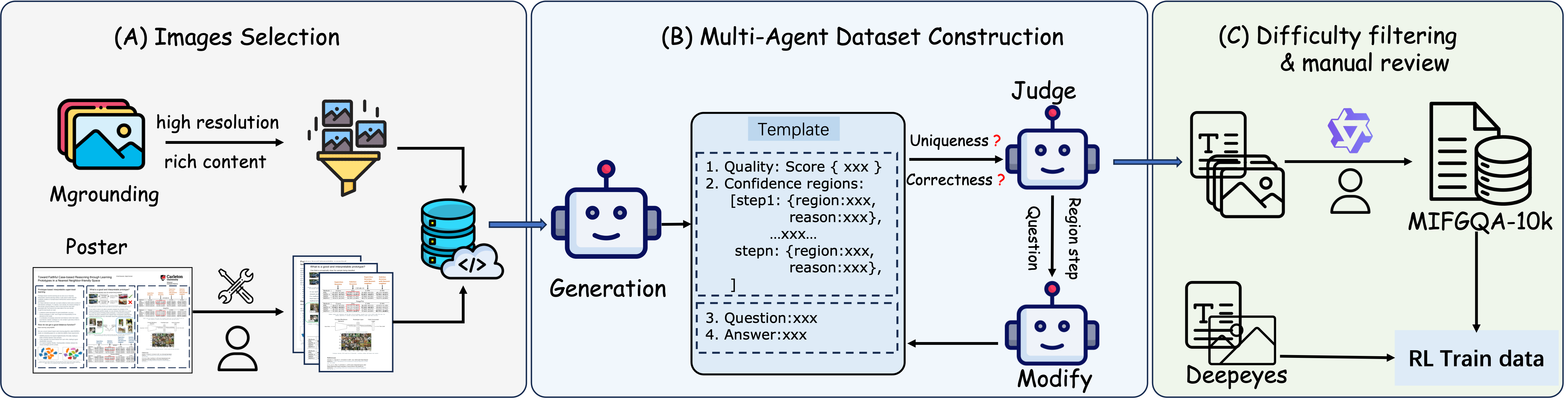}
   \caption{A three-stage data construction pipeline based on multi-agent systems.}
   \label{fig:data_construct}
\end{figure*}

\subsection{Reasoning Pattern}

Our goal is to train a vision agent capable of complex multi-image tasks through end2end reinforcement learning. Equipped with confirmation and reflection capabilities, \MODELNAME~can autonomously verify local details and revisit the global context, fostering a dynamic visual reasoning process. Even in late-stage inference, the agent maintains a high degree of attention to the image and acquires requisite information via attention refocusing driven by visual confirmation and attention redistribution driven by visual reflection.

Adhering to the ``observe--confirm/reflect--iterate'' paradigm, \MODELNAME~outputs the final answer after obtaining sufficient information. Specifically, the complete reasoning process can be described as follows:
\begin{enumerate}
    \item \textbf{Initial Observation.} Given an image set $I$ and a corresponding question $Q$, \MODELNAME~performs an initial reasoning step. During this stage, the model autonomously decides whether to invoke visual confirmation or reflection. If the model calls a tool, it will output structured tool use tags.

    \item \textbf{Calling Visual Tools.} When the model determines that there is a need for a visual tool call, it will generate the tool name and the required parameters.
    For \emph{visual confirmation} via local image cropping, the model outputs an image index to specify the target image, a bounding box to indicate the region to zoom in on, and a label to emphasize the region of interest. 
    % Upon successful execution, the cropped region is returned to the model.
    For \emph{visual reflection} via image reuse, the model outputs an image index and a reflection reason, which reinforces the purpose of reflection and enhances logical interpretability. 
    % After execution, the reflected image is returned to the model.
    \item \textbf{Observation and further reasoning.} The returned results are concatenated with the historical context in an appropriate format, forming the input for the current step. The model continues reasoning based on this augmented trajectory.

    \item \textbf{Iteration.} \MODELNAME~repeats Steps 2--3 until it either reaches the maximum interaction limit or produces an answer, at which point inference terminates.    
\end{enumerate}

\subsection{End2End Reinforcement Post-training}

We employ an end2end reinforcement learning framework to train the model. The core objective is to cultivate the model's dynamic visual reasoning ability—that is, to adaptively optimize tool invocation decisions based on multi-turn feedback, reallocate visual attention, and ultimately achieve a visual agent integrating multi-turn tool inference. We use the GRPO algorithm to optimize the parameterization strategy $\pi_\theta$, enabling the visual agent to maximize the expected cumulative reward of the task by dynamically adjusting the lexical generation logic, while removing KL regularization.

\subsubsection{Reward Shaping}

We design both result rewards and format rewards, with an additional tool use gain incorporated into the result rewards to incentivize the model's exploration and utilization of tools.
\begin{itemize}
\item \textbf{Result Reward}: Evaluated via rule-based judgment or an LLM judge, with a binary reward of 0 or 1.
\item \textbf{Format Reward}: Assesses whether the model output meets all specified criteria through rule checking (e.g., completeness of tags, absence of overlap between tags). It is designed as a gradient reward, where non-compliance results in corresponding score deductions.
\end{itemize}

The total reward is defined as:
$$R(\tau) = R_{\text{acc}} \cdot \left(a + b \cdot \text{Tool}_{\text{gain}}\right) + c \cdot R_{\text{format}}$$
where $\text{Tool}_{\text{gain}}$ represents a tool use gain (an indicator function for tool use) specifically introduced for the accuracy reward. This gain encourages the agent to obtain additional rewards through tool use under the Zero RL paradigm.

\subsubsection{Two-Level Mask Strategy}
To adapt to the training of visual tools, we design a two-level mask strategy to enhance training stability (as illustrated in Figure. \ref{fig:twomask}).

\textbf{Action-Level Mask.}

After the model issues a tool call request, the execution results of the tool (including the execution status and returned images) are concatenated into the previous context and used as input for the next round of reasoning. However, since this information often deviates from the distribution of existing knowledge, it can lead to abnormal directions in parameter optimization. To mitigate this issue, we apply a mask to information other than the model’s output to prevent it from participating in policy updates.

\textbf{Trajectory-Level Mask.}
During training, we observe that some interaction trajectories produce excessively long outputs or exceed the predefined maximum interaction rounds. These issues induce training instability and impede the model from learning effective tool-usage strategies. To address this problem, we propose a trajectory-level masking mechanism to filter out such invalid trajectories.

Specifically, we evaluate the $N$ trajectories generated for each sample in the GRPO algorithm. If a trajectory is identified as invalid, the entire trajectory is masked and excluded from policy updates. This mechanism introduces nearly no additional computational overhead, since it directly masks existing generated trajectories rather than generating new valid ones.
In this work, a trajectory is marked invalid if it exceeds the maximum interaction iterations, surpasses the maximum response length (typically caused by repetitive model outputs), or fails to produce any final answer. This mask also avoids reward-induced suppression of the model's long-range exploration.

\subsection{Curation of \DATASET}

To train the model's dynamic visual reasoning capabilities for multi-image tasks, we design a three-stage data synthesis pipeline (as shown in Figure~\ref{fig:data_construct}).
Stage 1: Images Selection.
Considering the necessity of activating visual capabilities, we exclude low-resolution images and selected data with rich visual content. Specifically, for natural images, we randomly sample high-resolution data from MGrounding~\cite{li2025migician}, covering image groups with relationships such as difference, contrast, and time.

For scientific figures, we constructe the dataset from high-resolution conference posters~\cite{saxena2025postersum}. We perform adaptive structural segmentation on individual conference posters, treating different segmented regions as multi-image data. The resulting multi-image dataset maintains strong contextual relevance.

Stage 2: Multi-Agent Data Construction.
First, a question generation agent is asked to generate fine-grained QA pair according to the given original images. To ensure the high quality and credibility of the generated questions, the Agent is also required to output structured reasoning steps from questions to answers, including confidence regions and reasoning details. Subsequently, an Answer Verification Agent evaluates the uniqueness of answers and the correctness of reasoning logic based on the questions and reasoning steps. If the criteria are not met, a Question Revision Agent is activated to iteratively refine the questions until the requirements are satisfied. 

Stage 3: Quality Filtering and Difficulty Calibration.
To ensure the training data has appropriate difficulty, We perform five rollouts on the entire dataset using Qwen2.5-VL-7B to select samples of moderate difficulty, removing samples that are too difficult or too simple. Then, we conduct further rule-based screening and invite several human experts to conduct manual quality checks to ensure the quality of the training data. More information on data construction is provided in the appendix.

We ultimately obtain approximately 10k samples, with an average total pixel count per question of 3 million (for natural images) and 9 million (for poster images). From this, we derive MIFG-QA-val, a benchmark for evaluating models' multi-image multi-hop fine-grained reasoning abilities. Combined with Deepeyes single-image data, this dataset serves as the training data for \MODELNAME.

\section{Experiments}
\subsection{Experimental Setup}

\begin{table*}[htbp]
\centering
\small
\caption{Results of different models on benchmarks. Bold text indicates the best score.}
\label{tab:model_results}
\resizebox{\linewidth}{!}{%
% 将 tabular 替换为 NiceTabular
\begin{tabular}{lccccccccccc}
\toprule
% 第一行表头：分组名称
\multirow{2}{*}{\textbf{Model}} & 
\multirow{2}{*}{\textbf{No SFT}} & 
\multicolumn{5}{c}{\textbf{Single-Image Benchmarks}} & 
\multicolumn{5}{c}{\textbf{Multi-Image Benchmarks}} \\ % 【修正】这里 4 改成了 5
\cmidrule(r){3-7} \cmidrule(l){8-12}
% 第二行表头：具体指标
& 
& 
\textbf{V*} & \textbf{HR-4K} & \textbf{HR-8K} & \textbf{MME} & \textbf{Avg} & 
\textbf{MIFG-QA} & \textbf{BLINK} & \textbf{Mantis} & \textbf{M4Bench} & \textbf{Avg} \\
\midrule

GPT-4o~\cite{hurst2024gpt4o} & -- & 66 & 65 & 59.6 & 62.8 & 63.4 & 48.11 & 60 & -- & -- & --  \\
Gemini2.5-Flash~\cite{comanici2025gemini} & -- & 72.3 & 77.5 & 73.7 & 60.9 & 71.1  & --  & --  & -- & -- & --  \\
Gemini2.5-pro~\cite{comanici2025gemini} & -- & 79.1 & 83.9 & 81.5 & 71.3 & 79.0  & -- & -- & -- & -- & -- \\ % 【修正】Gemoni 改为 Gemini
Qwen3-VL-235B~\cite{bai2025qwen3} & -- & 80.6 & 83.0  & 80.4 &  71.6 & 78.9  & -- & -- & -- & -- & -- \\
\midrule
Qwen2.5-VL-7B~\cite{bai2025qwen2} &  --  & 77.0 & 70.6 & 66.7 & 57.3 & 67.9 & 41.98 & 55.5 & 75.6 & 45.2 & 54.6 \\
Qwen2.5-VL-7B$_{\text{cot}}$ &  --  & 74.9 & 68.5 & 62.3 & 59.6 & 66.31 & 39.3 & 54.4 & 72.4 & 43.2 & 52.3 \\
SEAL~\cite{wu2024vstar} & -- & 75.4 & -- & -- & -- & -- & -- & -- & -- & -- & -- \\
Dyfo~\cite{li2025dyfo} & -- & 81.2 & -- & -- & -- & -- & -- & -- & -- & -- & -- \\
Deepeyes~\cite{zheng2025deepeyes} & \ding{52} & 85.6 & 75.0 & 69.8 & 64.1 & 73.6 & 38.36 & 55.7 & 75.1 & 54.6 & 56.1 \\
Pixel-Reasoner~\cite{su2025pixel}  & \ding{55} & 84.3 & 72.6 & 66.1 & 64.4 & 71.9 & 44.49 & 56.3 & 78.1 & 42.8 & 55.6 \\
Deepeyes-V2~\cite{hong2025deepeyesv2} & \ding{55} & 81.8 & \textbf{77.9} & \textbf{73.8} & 64.9 & 74.6 & -- & -- & -- & -- & -- \\
Thyme~\cite{zhang2025thyme} & \ding{55} & 82.2 & 77.0 & 72.0 & 64.8 & 74.0 & -- & 56.1 & -- & -- & -- \\
SenseNova-Mars-7B~\cite{chng2025sensenova} & \ding{55} & 79.1 & 69.1 & -- & -- & -- & -- & -- & -- & -- & -- \\
\MODELNAME-7B & \ding{52} & \textbf{88.5} &74.0  & 71.5  & \textbf{64.9} & \textbf{74.7}  &\textbf{49.21} &  \textbf{56.7} & \textbf{78.8} & \textbf{56.5}& \textbf{60.0} \\
\midrule
Qwen3-VL-8B~\cite{bai2025qwen3} & -- & 86.4 & 78.9 & 74.6 & 61.9 & 75.5 & 47.17 & 66.7 & 77.8 & 47.4 & 59.8 \\
Qwen3-VL-30B~\cite{bai2025qwen3}  & --  & 82.2 &  78.5 & 74.2 & 67.7 & 75.7 & 48.11 & 66.4 & 77.8 & 53.4 & 61.4 \\
Skywork-R1V4(30B)~\cite{zhang2025skywork} & \ding{55} & 88.0 & 82.8  & 79.8 & \textbf{71.4} & 80.3 & -- & -- & -- & -- & -- \\
SenseNova-Mars-8B~\cite{chng2025sensenova} & \ding{52} & \textbf{92.2} & 83.1 & 78.4 & 67.9 & 80.4 & 50.47 & 66.3 & 77.8 & 54.2 & 62.2 \\
\MODELNAME-8B& \ding{52} & 90.6 &\textbf{ 85.3 }& \textbf{82.9} & 70.5 &\textbf{ 82.3} &\textbf{ 53.86} &\textbf{ 67.8} & \textbf{78.3 }& \textbf{54.7} & \textbf{63.7} \\
\bottomrule
\end{tabular}
}
\end{table*}

\textbf{Hyperparameters.} We adopt an end2end reinforcement learning algorithm, GRPO, with Qwen3-VL-8B-Instruct as the base model. The average number of pixels per sample in our \DATASET~(poster) exceeds 9 million, which occupies over 12k context tokens. We therefore set the maximum pixel budget to 4 million. This preserves context space to accommodate more interaction turns. The model is trained on H20 GPUs with a batch size of 256, 8 rollouts per prompt, and a maximum limit of 5 tool interactions. The KL coefficient is set to 0.0, with the maximum input length restricted to 10,480 tokens and the maximum response length to 20,480 tokens. 
% Our single-image training data is sourced from Deepeyes, which includes fine-grained data of natural images and scientific figures, as well as reasoning data. To enable integrated reasoning over multi-image inputs, we incorporate the constructed \DATASET~dataset. 
Main experimental results are presented in Table~\ref{tab:model_results}, where the values are the average of repeated experiments, and the maximum pixel limit of input images is 16384×28×28.\\ 
\textbf{Baselines and Benchmarks.} We conduct a systematic comparison of \MODELNAME~against advanced proprietary models (e.g., GPT-4o~\cite{hurst2024gpt4o}), open-source models (e.g., Qwen2.5-VL~\cite{bai2025qwen2}), and models implementing multimodal CoT (e.g., Deepeyes~\cite{zheng2025deepeyes}, Pixels Reasoner~\cite{su2025pixel}, SenseNova~\cite{chng2025sensenova}). 

For evaluation benchmarks, we perform comprehensive comparisons on three classic single-image fine-grained benchmarks (V* Bench~\cite{wu2024vstar}, HR-Bench~\cite{wang2025hrbench}, and MME-RealWorld~\cite{zhang2024mme}) and four multi-image datasets (BLINK~\cite{fu2024blink}, Mantis~\cite{jiang2024mantis}, M4Bench~\cite{ye2025m4bench}, and \DATASET-val) to validate the effectiveness of \MODELNAME's dynamic visual reasoning.  The test data of V* Bench and HR-Bench includes ultra-high-resolution images with resolutions ranging from 2k to 8k, where the target objects relevant to the test questions account for a very small proportion of pixels. 
% Such extensive redundant visual information increases the difficulty for VLMs to accurately locate targets and answer questions. 
MME-RealWorld is a large-scale high-resolution dataset of real-world scenarios, covering 43 scenarios, which focuses on evaluating the perception and reasoning capabilities of multimodal large models. BLINK~\cite{fu2024blink} and Mantis~\cite{jiang2024mantis} are general multi-image benchmarks, while M4Bench~\cite{ye2025m4bench} and \DATASET-val are benchmarks that focus on evaluating fine-grained understanding.

% \vspace{-5pt}
\subsection{Main Results}

In Table~\ref{tab:model_results} of the experimental results, we reveal a noteworthy phenomenon that, for base Qwen2.5-VL-7B, on V* and HR-Bench, directly generated answer is superior to the step-by-step generated ones, especially on the higher resolution tasks(HR-Bench 8K). 
On our MIFG-QA, where the input images have relatively high total resolution and abundant fine-grained information, we observe a similar phenomenon. We speculate that this phenomenon stems from the issue revealed in Figure~\ref{fig:attention-decay}.

While existing methods achieve significant improvements in single-image scenarios through external visual tools, they largely overlook multi-image tasks—either lacking native support or failing to evaluate in such settings. To benchmark these vision agents on multi-image data, we modify prompts for models incompatible with multi-image operations, forcing them to output image indices during tool call. However, some models, such as the DeepeyesV2 configuration code tool, are fundamentally unable to handle multiple image inputs.

\begin{table}[htbp]
\centering
\vspace{-5pt} % 缩小表格与上段文字距离
\caption{Average Tool Call Count by Benchmark. Values in superscript indicate (zoom in ratio / reuse ratio)}
\label{tab:zoom_reuse_comparison}
% \vspace{-8pt} % 缩小标题与表格距离
\resizebox{\linewidth}{!}{%
\begin{tabular}{cccc}
\toprule
V*  & HR-Bench & MME & MIFGQA \\
\midrule
1.00\textsuperscript{~(1.00~/~0.00)} &
1.05\textsuperscript{~(0.97~/~0.03)} &
1.12\textsuperscript{~(0.93~/~0.07)} &
1.70\textsuperscript{~(0.70~/~0.30)} \\
\bottomrule
\end{tabular}
}
\end{table}
\vspace{-4pt} % 缩小表格与上段文字距离

\begin{table*}[htbp]
\centering
\caption{Ablation experiments: ``text RL" means performing RL training on plain-text CoT without any tools,``w/o MIFG-QA" refers to adopting only single-image training data, ``w/o lookback" denotes not using visual reflection tools, and ``w/o Traj mask" signifies not masking invalid tracks.}
\label{tab:ablation}
\resizebox{\linewidth}{!}{%

\begin{tabular}{lcccccccccccc}
\toprule
\multirow{3}{*}{\textbf{Components}} &
\multicolumn{3}{c}{\textbf{V*}} &
\multicolumn{2}{c}{\textbf{HR-Bench}} &
\multirow{3}{*}{\textbf{MME}} &
\multicolumn{3}{c}{\textbf{MIFG-QA}} &\multirow{2}{*}{\textbf{ BLINK}} & \multirow{2}{*}{\textbf{Mantis}} &\multirow{2}{*}{\textbf{ M4Bench}} \\
\cmidrule(lr){2-4}\cmidrule(lr){5-6}\cmidrule(lr){8-10}
 & 
\textbf{direct} & \textbf{relative} & \textbf{Avg} &
\textbf{4K} & \textbf{8K}  &
 &
\textbf{nature} & \textbf{poster} & \textbf{Avg} & & & \\
\midrule
Baseline & 79.13 & 73.68 & 76.96 & 70.62 & 66.75 & 57.3 & 36.94 & 44.34 & 41.98 & 55.5 & 75.6 & 45.2 \\
Baseline+text RL & 82.61 & 78.95  & 81.15 & 72.75 & 68.75 & 63.57 & 45.81 & 42.26 & 43.40 & 55.2 & 77.4 & 48.9 \\
\MODELNAME-7B w/o Tool gain &  86.08 & 82.89 & 84.81 & 72.37 & 71.0 & 63.14 & 46.80 & 44.34  & 45.34 & 55.1 & 78.3 & 47.6  \\
\MODELNAME-7B w/o MIFG-QA & 87.83 & 80.26 & 84.81 & 72.75 & 70.5   & 64.02 & 39.90 & 40.18 & 40.09 & 56.1 & 77.0 & 53.5 \\
\MODELNAME-7B w/o lookback & 86.07 & 76.32 & 82.20 & 74.0  & 71.25 & 62.24 & 48.25 & 47.98 & 48.07 & 56.0 & 78.3 & 53.9\\
\MODELNAME-7B w/o Traj mask & 86.07 & 81.59 & 84.29 & 73.5  & 69.75 & \textbf{}{64.03} & 46.31 & 46.18 & 46.23 & \textbf{56.7 }& 78.3 & 54.5\\
\MODELNAME-7B & \textbf{88.70 }& \textbf{88.16} & \textbf{88.48} & \textbf{74.0}  & \textbf{71.5}    &\textbf{ 64.90} & \textbf{50.24} & \textbf{48.73} & \textbf{49.21} & 56.4 & \textbf{78.8} &\textbf{55.6} \\
\bottomrule
\end{tabular}
}
\end{table*}

\MODELNAME~Utilizes visual reflection and confirmation mechanisms to autonomously think about key images/regions, achieving state-of-the-art performance on single-image and multi-image tasks, and showing significant improvements on fine-grained tasks (V*, HR-Bench, MIFG-QA, M4Bench, etc.). In particular, we present tool usage statistics for the evaluation of \MODELNAME~in the Table~\ref{tab:zoom_reuse_comparison}. For multi-image data, the proportion of visual reflection tool use increases significantly, indicating that the model tends to review images to avoid forgetting, thereby achieving more accurate reasoning in multi-image scenarios.
Notably,  \MODELNAME~acquires dynamic visual reasoning capabilities and exhibits diverse tool-usage strategies through reinforcement learning alone. Representative cases include direct or exploratory visual confirmation, cross-image comparison, and cross-image reflective confirmation. More detailed reasoning examples are provided in the supplementary materials.

We also conducted inference experiments under limited-resolution scenarios, and the results are shown in Table~\ref{tab:pixels}. Experiments demonstrate that when the maximum input resolution decreases, the problem becomes more difficult.
All models exhibit a downward trend in overall performance; however, models with multimodal CoT still retain their performance advantages, being significantly superior to base models. Furthermore, our model still performs excellently under constrained scenarios, which demonstrates the adaptability of \MODELNAME's dynamic visual thinking.

\begin{table}[htbp]
  \centering
  \caption{Model Performance Comparison Across Different MaxPixels. MP means maxpixels of input image.}
\resizebox{\linewidth}{!}{%
  \begin{tabular}{clcccc}
    \toprule
    \multirow{2}{*}{\textbf{MP}} & \multirow{2}{*}{\textbf{Model}} & \multirow{2}{*}{\textbf{V*}} & \multicolumn{2}{c}{\textbf{HR-Bench}} & \multirow{2}{*}{\textbf{MME}} \\
    \cmidrule(lr){4-5}  % 仅在hebench-4子列下画横线
    & & & \textbf{4K} & \textbf{8K} & \\
    \midrule
    \multirow{4}{*}{12M} & Qwen2.5-VL & 76.96 & 70.62 & 66.75 & 57.3 \\
    & Pixel Reasoner & \underline{85.86} & 68.75 & 63.25 &\underline{ 64.41} \\
    & Deepeyes & 85.56 & \textbf{75.0} & \underline{69.75} & 64.0 \\
    & IMAgent-7B & \textbf{88.48} & \underline{74.0} & \textbf{71.5} & \textbf{64.90} \\
    \midrule
    \multirow{4}{*}{4M} & Qwen2.5-VL & 75.39 & 69.75 & 64.25 & 56.76 \\
    & Pixel Reasoner & 85.34 & 68.75 & 62.25 & \underline{63.91} \\
    & Deepeyes & \underline{85.56} & \textbf{73.0} &\textbf{ 67.75} & 63.86 \\
    & IMAgent-7B &\textbf{ 88.48} & \underline{71.5} & \underline{66.12} & \textbf{64.33} \\
    \midrule
    \multirow{4}{*}{1M} & Qwen2.5-VL & 71.73 & 62.38 & 52.12 & 51.35 \\
    & Pixel Reasoner & 77.49 & \underline{65.25} & 55.12 & \underline{61.51} \\
    & Deepeyes & \underline{80.10} & \textbf{65.75} & \underline{55.25} & 60.23 \\
    & IMAgent-7B & \textbf{80.62} & 64.12 &\textbf{ 55.62} &\textbf{ 61.67} \\
    \bottomrule
  \end{tabular}
  }
  \label{tab:pixels}
\end{table}

\subsection{Ablation Study}
\begin{figure}[htbp]
  \centering
   \includegraphics[width=0.98\linewidth]{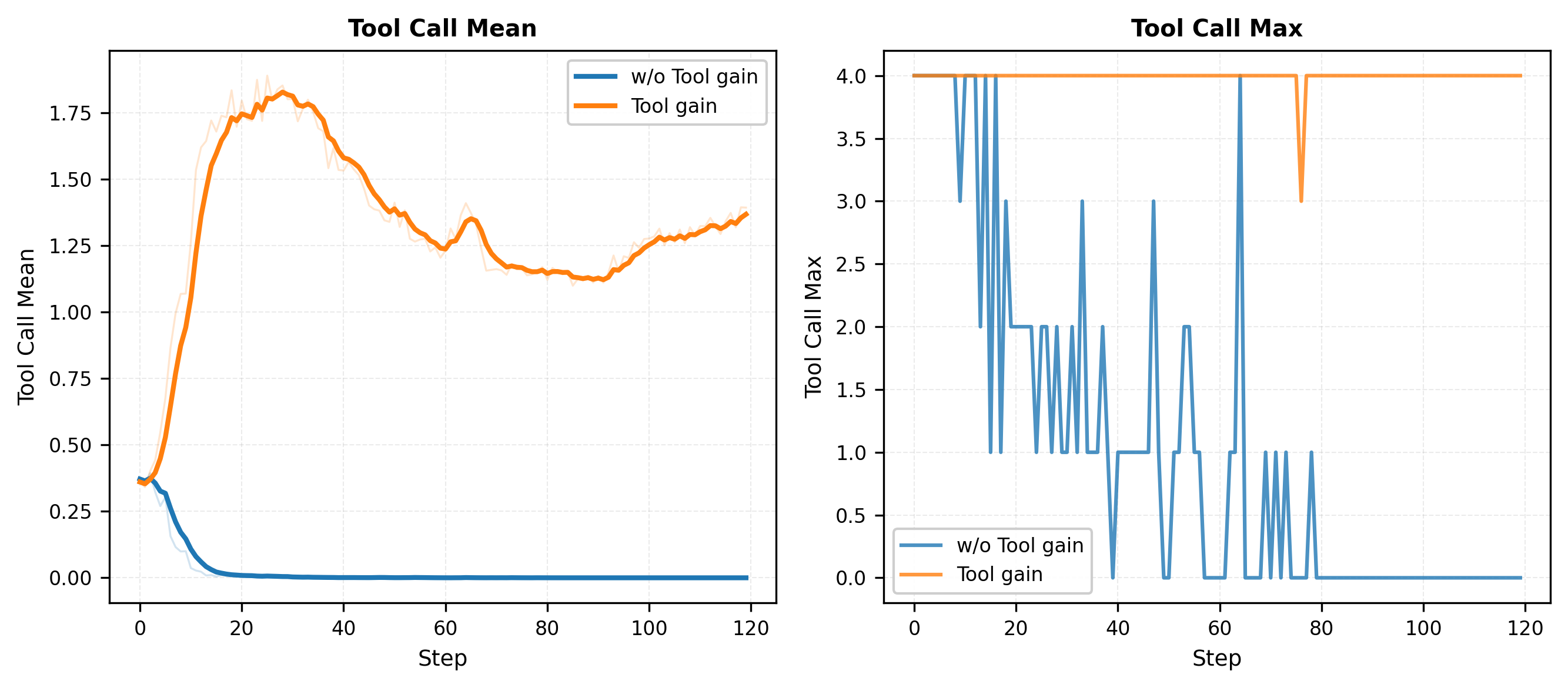}
   \caption{Comparison of tools with and without tool gain.}
   \label{fig:tool_call_comparison}
\end{figure}

We present the training curves of the average tool use counts for models with and without tool gain, as shown in Figure \ref{fig:tool_call_comparison}. When the tool gain is set to zero, the model's exploratory utilization of tools gradually decays to zero as training steps increase. This phenomenon is common in non-native agentic models such as Qwen2.5-VL-Instruct (while not in Qwen3-VL). The pre-training corpus of these models lacks interactive data related to tool use, resulting in weak tool interaction capabilities. After introducing the accuracy-based tool use gain, the model is incentivized to learn strategies that leverage tools to derive correct answers while avoiding meaningless tool use. As training proceeds, the average number of tool use instances per step for \MODELNAME~increases gradually and peaks in the early training stage—demonstrating that the tool gain effectively encourages the model to explore tool use during the initial phase. In the mid-training stage, the average per-step tool use count decreases steadily, indicating that the model has adapted to completing visual reasoning tasks with tool assistance and evolved toward more efficient tool use strategies. In the late training stage, the model masters diverse reasoning strategies integrated with tool use, and the average per-step tool use count tends to stabilize.

We further investigate the effects of visual tools, training data, and trajectory-level mask on the ability of Qwen2.5-VL-7B (which lacks native tool-usage capability) to learn dynamic visual skills, so as to verify the robustness of each component. The results are presented in Table \ref{tab:ablation}.

A comparison between Row 5 and Row 7 demonstrates that when equipped with the visual reflection tool, the model’s overall performance (on the V*-relative, MME, MIFG-QA and M4Bench datasets) is significantly improved. Explicit visual reflection enables the model to re-examine images, thereby achieving more precise alignment between questions and image regions. For multi-image scenarios, the reflection tool assists the model in identifying images relevant to the query.

Experiments in Row 4 and Row 7 show that incorporating multi-image data into the training process helps comprehensively enhance model performance. Multi-image data involves more image elements and comparisons between images, which effectively expands the diversity of tasks the model learns and encourages the model to explore richer tool use strategies.
A comparison between RoW 6 and Row 7 indicates that the model’s multi-image processing capability drops sharply if trajectory-level mask is not applied during training. Multi-image tasks typically require more visual operations and interaction rounds. The lack of masking for invalid trajectories inhibits long-range interactions, thus limiting the model's exploration of tool-use strategies for multi-image problems during training.

In summary, the multi-image training data we add helps the model learn tool use methods in multi-image scenarios and improves its overall performance. The proposed visual reflection function allows the model to re-evaluate images. Filtering invalid trajectories expands the model’s tool exploration space and enhances its multi-image processing capability. The synergistic effect of these three components endows the model with dynamic visual thinking capabilities and boosts its overall performance.

\section{Dynamic Visual Attention in Reasoning}
\subsection{Visual Tools Enable Attention Reallocation}

\begin{figure}[htbp]
  \centering
   \includegraphics[width=0.98\linewidth]{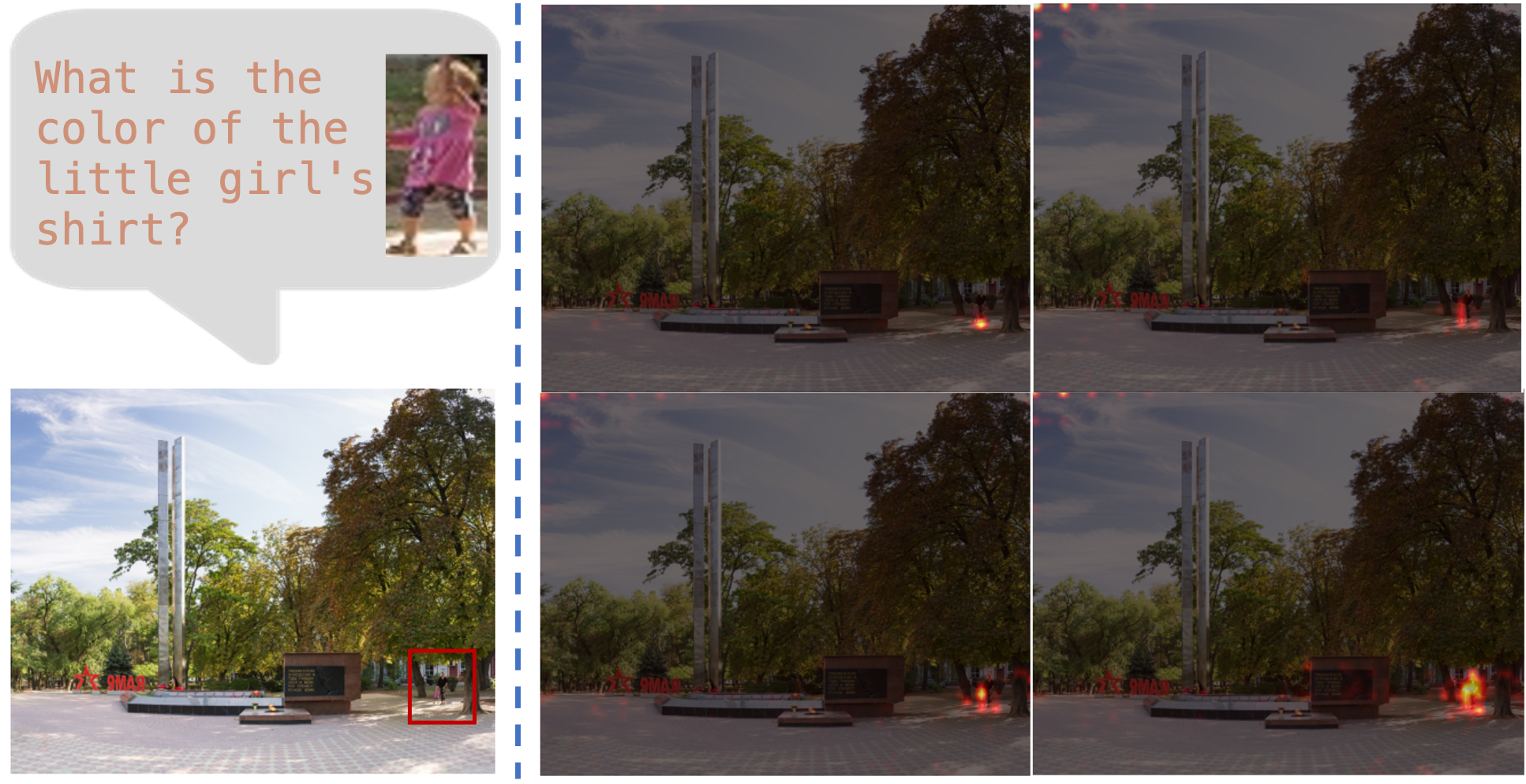}

   \caption{Attention maps for the visual confirmation tool. Left of the dashed line lies the query image pair; the text box thumbnail marks the cropped region. Right of the line are attention maps for one token each when the model outputs four bounding boxes.}

% 图像使用视觉确认工具的注意力热力图。蓝色虚线左侧是问题图像对，其中1文本框中的小图是模型使用视觉确认工具的裁剪区域，虚线右侧是模型输出四个边界框时各取一个token对应的注意力热图。
   \label{fig:bodingbox_attention}
\end{figure}
Figure~\ref{fig:attention-decay}(a,b) show that the general-purpose vision model Qwen2.5-VL gradually stops attending to images in later reasoning steps, especially in multi-image tasks.
The model ``sees" the images only at the initial reasoning stage; thereafter, even when the reasoning involves image content, it fails to enhance attention focus on the corresponding image regions. This reliance solely on the visual ``first impression" stems from the model’s training bias toward continuing reasoning based on already generated linguistic content rather than revisiting visual evidence. When facing complex visual scenarios, this often leads to performance degradation.

In contrast, experiments in Figure~\ref{fig:attention-decay} (c, d) demonstrate that \MODELNAME, equipped with confirmation and reflection tools, can proactively revisit visual evidence. This enables it to maintain autonomous visual attention allocation throughout the entire reasoning process and achieve visual focus on key regions. 
\begin{figure}[htbp]
  \centering
   \includegraphics[width=0.95\linewidth]{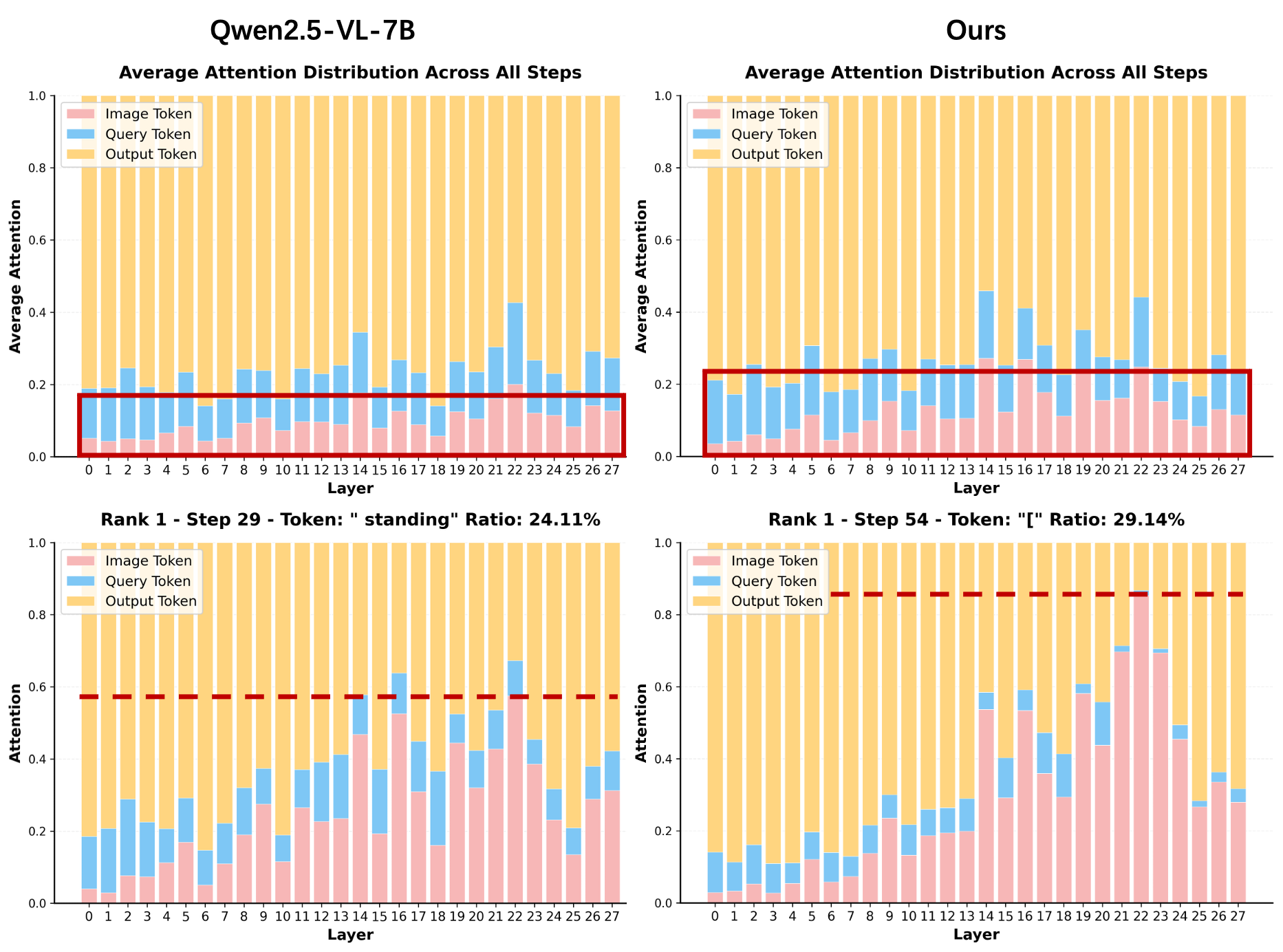}
   \caption{Attention distribution across model layers.}
% 图像各层注意力分布图。第一行是模型所有输出token对三类token的注意力比例在各层的分布，第二行是模型对图像注意达到峰值的输出token的各层注意力分布
   \label{fig:layer_atten_ratio}
\end{figure}
Additionally, we observe that the model pays higher attention to images when calling tools. These moments of heightened attention occur when the model outputs bounding boxes for the regions to be confirmed— corresponding to the shaded `` [bounding box] " segments marked in Figure~\ref{fig:attention-decay} (c). Figure~\ref{fig:bodingbox_attention} show that \MODELNAME~ highly focuses on the problem-related regions during bounding box output. This enhanced attention to images is primarily concentrated on image regions relevant to the problem, rather than being equally or similarly elevated across all regions.

We further analyze the attention distribution across all layers of the model, as shown in Figure~\ref{fig:layer_atten_ratio}. The first row shows the layer-wise attention distribution of all output tokens to the three token categories. The second row presents the layer-wise attention distribution of the output tokens where the model’s attention to images reaches its peak. We found that higher image attention typically appears in the middle and later layers of the model, and \MODELNAME~also exhibits significantly higher image attention in these layers. This phenomenon is more pronounced in the comparison of attention distribution at the token level: \MODELNAME~demonstrates far greater visual attention than Qwen2.5-VL across multiple intermediate layers.

\subsection{Visual Tools Drive Attention Focus on Task-Relevant Regions}

\begin{figure}[htbp]
  \centering
   \includegraphics[width=0.98\linewidth]{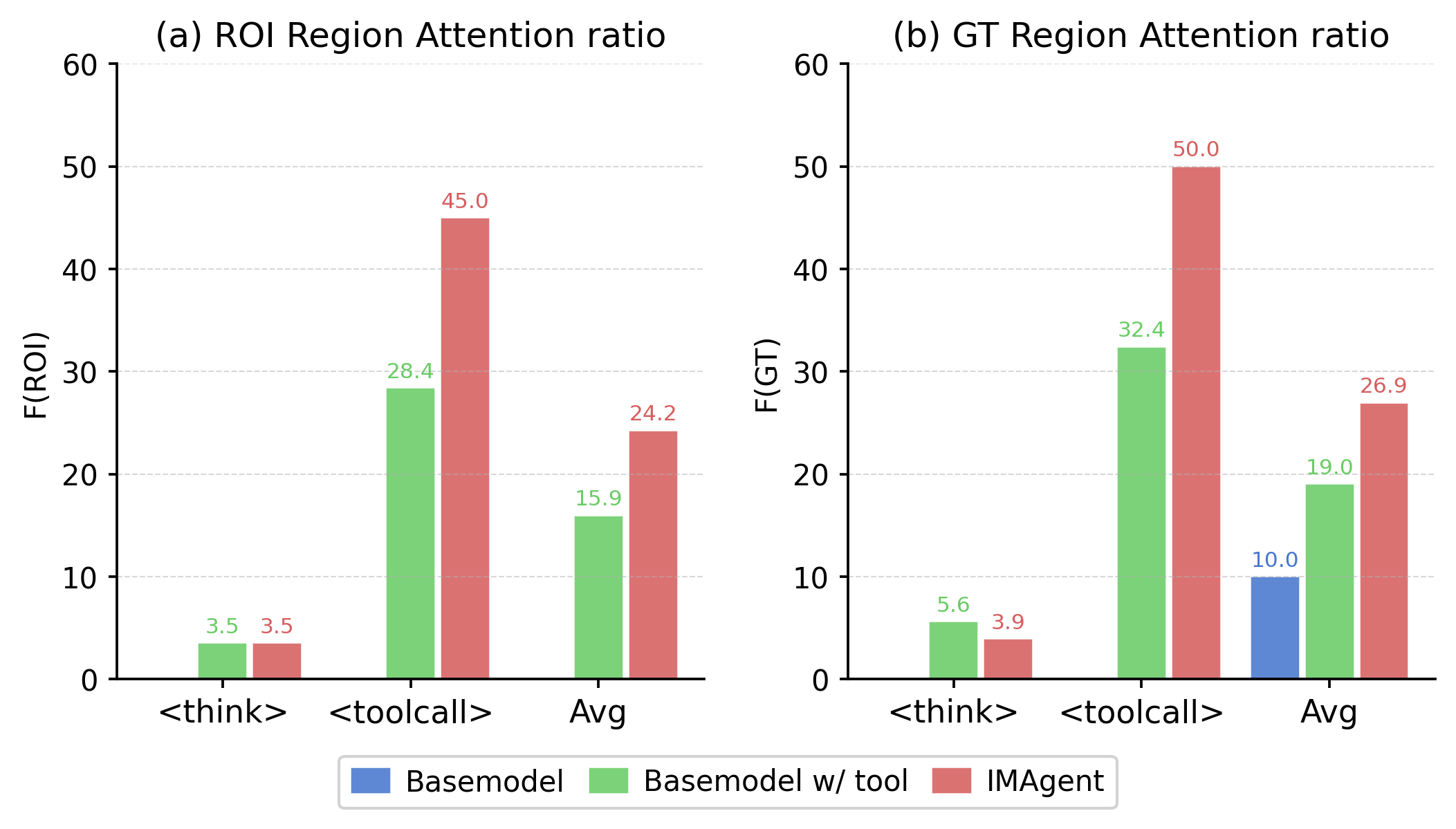}
   % \caption{Results of the average attention ratio across different visual regions. Here, the ROI regions refer to the bounding box regions of tool calls, and the GT regions refer to the original answer-related regions in the dataset.}
   \caption{Results of average attention ratio across visual regions. ROI denotes tool-call bounding boxes region; GT indicates original answer-related dataset regions.}
   \label{fig:model_atten_scomparison}
\end{figure}

To further validate the above observations quantitatively, we evaluate on 100 samples from the V* dataset~\cite{wu2024vstar}.
We define an attention ratio score:$F(\mathcal{X}) = \frac{\text{Mean}(\text{Attention}(\mathcal{X}))}{\text{Mean}(\text{Attention}(\overline{\mathcal{X}}))}$, where the numerator and denominator denote the mean attention within region $\mathcal{X}$ and outside it, respectively.
A higher $F(\mathcal{X})$ indicates that the model attends more to region $\mathcal{X}$ relative to the rest of the image.
We compute $F(\text{ROI})$ and $F(\text{GT})$ to measure the model's focus on the tool-queried bounding box region and the answer-relevant region, respectively.
Results are shown in Figure~\ref{fig:model_atten_scomparison}.

Figure~\ref{fig:model_atten_scomparison} (a) shows that in the tool use setting, $F(\text{ROI})$ during the \texttt{<tool\_call>} phase is substantially higher than during the \texttt{<think>} phase.
This indicates a natural \emph{decoupling} between perception and reasoning: the model leverages tool calls for targeted visual perception while using the think phase for language-level reasoning.

Figure~\ref{fig:model_atten_scomparison} (b) reveals that the elevated $F(\text{GT})$ during \texttt{<tool\_call>} confirms that the ROI regions queried by tool calls are highly correlated with the answer-relevant areas.
This suggests that the performance gains of \MODELNAME~stem from more precise visual localization of task-relevant regions.
% And the changes in the  \texttt{<Avg>} score indicates that explicit tool calling can strengthen a model's visual attention toward answer-relevant regions, and that this effect can be further amplified through training.
Regarding the \texttt{<Avg>} score in (b), the baseline Qwen2.5-VL achieves $F(\text{GT})=10$ without tool use. Equipping the same model with tools raises this to $19.0$, while \MODELNAME~further improves it to $26.9$. This demonstrates that explicit tool calling can strengthen a model's visual attention toward answer-relevant regions, and that this effect can be further amplified through training.

Overall, \MODELNAME~consistently achieves higher attention ratios across both ROI and GT regions, with the most pronounced improvement in \texttt{<tool\_call>} tokens.
These results confirm that \MODELNAME~actively redirects visual attention to task-relevant regions throughout inference, rather than relying on a fixed first impression of the image.

\section{Conclusion}

In this work, we present \MODELNAME, an open-source vision agent trained via pure end2end reinforcement learning to address complex single-image and multi-image tasks.. By introducing a curated multi-image QA dataset, \DATASET, and equipping the agent with specialized tools for visual reflection and confirmation, our approach enables dynamic adjustment of image attention and stable tool use throughout reasoning. Extensive experiments show that \MODELNAME~achieves state-of-the-art performance on both single-image and multi-image benchmarks, offering valuable insights and practical guidelines for advancing the development of vision agents in real-world scenarios.

{
    \small
    \bibliographystyle{ieeenat_fullname}
    \bibliography{main}
}


\end{document}